# Look, Imagine and Match:
# Improving Textual-Visual Cross-Modal Retrieval with Generative Models

Jiuxiang Gu[1], Jianfei Cai[2], Shafiq Joty[2], Li Niu[3], Gang Wang[4]
[1] ROSE Lab, Interdisciplinary Graduate School, Nanyang Technological University, Singapore
[2] School of Computer Science and Engineering, Nanyang Technological University, Singapore
[3] Rice University, USA  [4] Alibaba AI Labs, Hangzhou, China
{jgu004, asjfcai, srjoty}@ntu.edu.sg, {ustcnewly, gangwang6}@gmail.com

## Abstract

*Textual-visual cross-modal retrieval has been a hot research topic in both computer vision and natural language processing communities. Learning appropriate representations for multi-modal data is crucial for the cross-modal retrieval performance. Unlike existing image-text retrieval approaches that embed image-text pairs as single feature vectors in a common representational space, we propose to incorporate generative processes into the cross-modal feature embedding, through which we are able to learn not only the global abstract features but also the local grounded features. Extensive experiments show that our framework can well match images and sentences with complex content, and achieve the state-of-the-art cross-modal retrieval results on MSCOCO dataset.*

## 1. Introduction

As we are entering the era of big data, data from different modalities such as text, image, and video are growing at an unprecedented rate. Such multi-modal data exhibit heterogeneous properties, making it difficult for users to search information of interest effectively and efficiently. This paper focuses on the problem in multi-modal information retrieval, which is to retrieve the images (resp. texts) that are relevant to a given textual (resp. image) query. The fundamental challenge in cross-modal retrieval lies in the heterogeneity of different modalities of data. Thus, the learning of a common representation shared by data with different modalities plays the key role in cross-modal retrieval.

In recent years, a great deal of research has been devoted to bridge the heterogeneity gap between different modalities [15, 12, 16, 8, 38, 7, 6, 3, 34]. For textural-visual cross-modal embedding, the common way is to first encode individual modalities into their respective features, and then map them into a common semantic space, which is often optimized via a ranking loss that encourages the similarity of the mapped features of ground-truth image-text pairs to be greater than that of any other negative pair. Once the common representation is obtained, the relevance / similarity between the two modalities can be easily measured by computing the distance (*e.g.* $l_2$) between their representations in the common space.

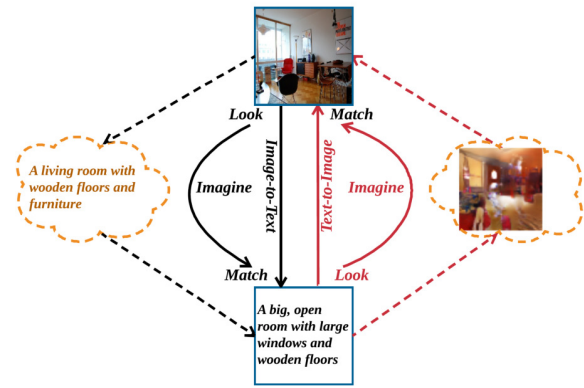

Figure 1: Conceptual illustration of our proposed cross-modal feature embedding with generative models. The cross-modal retrievals (*Image-to-Text* and *Text-to-Image*) are shown in different color. The two blue boxes are cross-modal data, and the generated data are shown in two dashed yellow clouds.

Although the feature representations in the learned common representation space have been successfully used to describe high-level semantic concepts of multi-modal data, they are not sufficient to retrieve images with detailed local similarity (*e.g.*, spatial layout) or sentences with word-level similarity. In contrast, as humans, we can relate a textual (resp. image) query to relevant images (resp. texts) more accurately, if we pay more attention to the finer details of the images (resp. texts). In other words, if we can ground the representation of one modality to the objects in the other modality, we can learn a better mapping.

Inspired by this concept, in this paper we propose to incorporate generative models into textual-visual feature embedding for cross-modal retrieval. In particular, in addition to the conventional cross-modal feature embedding at the global semantic level, we also introduce an additional cross-modal feature embedding at the local level, which is *grounded* by two generative models: image-to-text and text-to-image. Figure 1 illustrates the concept of our proposed cross-modal feature embedding with generative models at high level, which includes three learning steps: *look*, *imagine*, and *match*. Given a query in image or text, we first *look* at the query to extract an *abstract* representation. Then, we *imagine* what the target item (text or image) in the other modality should look like, and get a more concrete *grounded* representation. We accomplish this by asking the representation of one modality (to be estimated) to generate the item in the other modality, and comparing the generated items with gold standards. After that, we *match* the right image-text pairs using the relevance score which is calculated based on a combination of *grounded* and *abstract* representations.

The contributions of this paper are twofold. First, we incorporate two generative models into the conventional textual-visual feature embedding, which is able to learn concrete *grounded* representations that capture the detailed similarity between the two modalities. Second, we conduct extensive experimentations on the benchmark dataset, MSCOCO. Our empirical results demonstrate that the combination of the *grounded* and the *abstract* representations can significantly improve the state-of-the-art performance on cross-modal image-caption retrieval.

## 2. Related Works

Our work is closely related to the existing works on supervised cross-modal feature learning/embedding for cross image-text applications such as image captioning and image-text cross-modal retrieval. Particularly, the pairwise ranking is often adopted to utilize similar or dissimilar cross-modal data pairs to learn a proper similarity or distance metric between different modalities [36, 2, 24, 10].

Frome *et al*. [5] proposed a cross-modal feature embedding framework that use CNN and Skip-Gram [21] to extract cross-modal feature representations, and then associated them with a structured objective in which the distance between the matched image-caption pair is smaller than that between the mismatched pair. A similar framework is proposed by Kiros *et al*. [15], in which a Gated Recurrent Unit (GRU) was used as the sentence encoder. They also mapped the images and sentences to a common space and adopted the rank loss to penalize the model by averaging the individual violations across the negatives. Vendrov *et al*. [38] introduced an improved objective, which can preserve the partial order structure of a visual-semantic hierarchy. Klein *et al*. [17] adopted a similar objective and employed Fisher Vectors (FV) [29] as a pooling strategy of word embeddings for caption representation. In [18], they sold the visual word embedding idea and proposed a joint image-caption embedding model for image captioning. However, their model is based on cartoon-like images, which is difficult to be applied to real images. Considering the strong ability of Generative Adversarial Networks (GANs) in learning discriminative representation, Peng *et al*. [28] explored inter-modality and intra-modality with a cross-modal GAN. Recently, Faghri *et al*. [4] improved Kiros's work by replacing the sum violations across the negative samples with the hardest negative samples.

Several works have explored the alignment of visual objects and textual words [13, 30, 11, 25]. Karpathy *et al*. [13] used local alignment to embed the fragments of images and the sentences into a common space. Plummer *et al*. [30] went a step further and used all pairwise instances for similarity measurement. Jiang *et al*. [11] learned a multi-modal embedding by optimizing pairwise ranking, while enhancing both local alignment and global alignment. In [10], they introduced the context-modulated attention mechanism into the cross-modal embedding. Their attention scheme can selectively attend to pairwise instances of image and sentence, and then dynamically aggregate the measured similarity to obtain a global similarity between image and text. Instead of embedding the sentence with chain-structured RNNs, the recent work of Niu *et al*. [25] adopted a tree-structured LSTM to learn the hierarchical relations between sentences and images, and between phrases and visual objects. However, to align visual objects with textual words, a sufficient amount of annotations need to be acquired as well, which induces expensive human annotations.

Most of the existing studies on cross-modal textual-visual retrieval mainly focus on learning a high-level common space with ranking loss. In contrast, our approach learns not only the high-level global common space but also the local common space through generative models.

## 3. Proposed Generative Cross-modal Learning Network

### 3.1. System Overview

Figure 2 shows the overall architecture for the proposed generative cross-modal feature learning framework, named GXN. The entire system consists of three training paths: multi-modal feature embedding (the entire upper part), image-to-text generative feature learning (the blue path), and text-to-image generative adversarial feature learning (the green path). The first path is similar to the existing cross-modal feature embedding that maps different modality features into a common space. However, the difference here is that we use two branches of feature embedding, *i.e.*, making the embedded visual feature $v_h$ (resp. $v_l$) and the textual feature $t_h$ (resp. $t_l$) closer. We consider $(v_h, t_h)$ as high-level abstract features and $(v_l, t_l)$ as detailed grounded features. The grounded features will be used and regularized in the other two generative feature learning paths. The entire first training path mainly includes one image encoder $\text{CNN}_{\text{Enc}}$ and two sentence encoders $\text{RNN}_{\text{Enc}}^h$ and $\text{RNN}_{\text{Enc}}^l$.

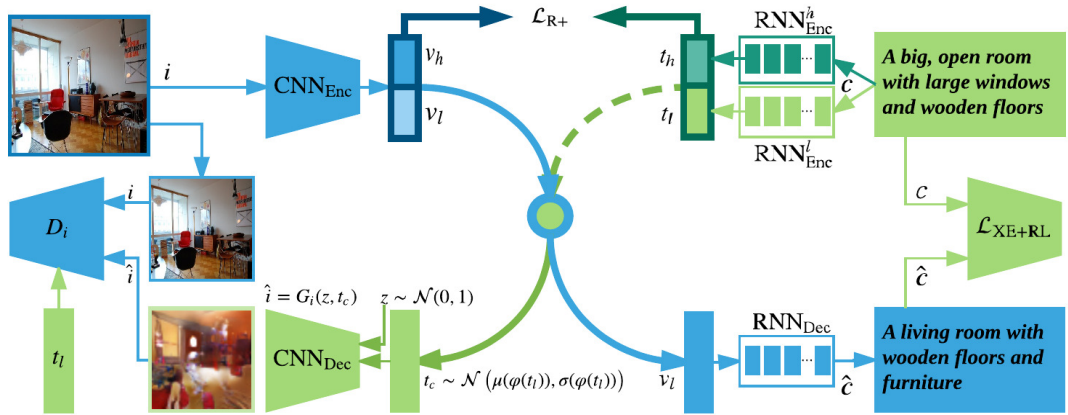

Figure 2: The proposed generative cross-modal learning framework (GXN). The entire framework consists of three training paths: cross-modal feature embedding (the entire upper part), image-to-text generative feature learning (the blue path), and text-to-image generative adversarial feature learning (the green path). It includes six networks: two sentence encoders $\text{RNN}_{\text{Enc}}^h$ (dark green) and $\text{RNN}_{\text{Enc}}^l$ (light green), one image encoder $\text{CNN}_{\text{Enc}}$ (blue), one sentence decoder $\text{RNN}_{\text{Dec}}$, one image decoder $\text{CNN}_{\text{Dec}}$ and one discriminator $D_i$.

The second training path (the blue path) is to generate a sentence from the embedded generative visual feature $v_l$. It consists of the image encoder $\text{CNN}_{\text{Enc}}$ and a sentence detector $\text{RNN}_{\text{Dec}}$. With a proper loss against ground-truth sentences, the grounded feature $v_l$ will be adjusted via back propagation. The third training path (the green path) is to generate an image from the textual feature $t_l$. Here we adopt the generative adversarial model, which comprises a generator / decoder $\text{CNN}_{\text{Dec}}$ and a discriminator $D_i$.

Overall, through these two paths of cross-modal generative feature learning, we hope to learn powerful cross-modal feature representations. During the testing stage, $\{v_h, v_l\}$ and $\{t_h, t_l\}$ will be used as the final feature representations for cross-modal retrieval, although the proposed GXN also produces other byproducts such as image-to-text generation and text-to-image generation, which are not the main focus of this paper. In the following, we describe each of the three training paths in detail.

### 3.2. Cross-modal Feature Embedding

We follow the common cross-modal feature embedding approach to embed the representations of the image and the caption into a common space, and then use a pairwise ranking loss to learn the model parameters [38]. In particular, given an image-caption pair $(i, c)$, where $i$ is the image and $c = (w_0, \cdots, w_{T-1})$ is the corresponding description with $w_i$ being the one-hot word encoding, we encode a caption by embedding each word in $c$ into a distributed representation using $\mathbf{W}_e w_i$, where $\mathbf{W}_e$ is a shared word embedding matrix to be learned. $\mathbf{W}_e$ can be initialized randomly or using pre-trained embeddings like word2vec [22]. Then we use two sequential sentence encoders (*e.g.*, GRU) to get the sentence representations. As for image encoding, we use a CNN that is pre-trained on ImageNet. More formally, we formulate the embedding and mapping of each modality as:

$$\begin{aligned} v_k &= P_v^k(\text{CNN}_{\text{Enc}}(i; \theta_i)) \\ t_k &= P_t^k(\text{RNN}_{\text{Enc}}^k(c; \theta_c^k)) \end{aligned}, \quad k \in \{h, l\} \quad (1)$$

where $\theta_i$ and $\theta_c^k$ are the parameters of the image and caption encoders, $P_v^k$ and $P_t^k$ are the linear transformation functions which map the encoded vectors into a common embedding space, and $v_k$ and $t_k$ are the resulting mapped vectors for the image and the caption, respectively.

We first consider the same pairwise ranking loss proposed in [15, 36, 12, 38]. We refer $(i, c)$ as positive pairs and denote the negative samples by $i'$ and $c'$, where $i'$ goes over images not described by $c$ and $c'$ goes over captions that do not describe $i$. We want the objective function to encourage the similarity of ground truth caption-image pairs to be greater than that of all other negative pairs. We, therefore, optimize the ranking loss of $\mathcal{L}_{\text{Rank}} = \frac{1}{N} \sum_{n=1}^{N} \mathcal{L}_{\text{R}}(i_n, c_n)$,

where the single sample ranking loss $\mathcal{L}_\text{R}$ is defined as:

$$\mathcal{L}_\text{R} = \sum_{t'} [\alpha - s(t,v) + s(t',v)]_+ + \sum_{v'} [\alpha - s(t,v) + s(t,v')]_+ \quad (2)$$

where $\alpha$ is a margin, $s(t,v) = -\|\max(0, v-t)\|^2$ is the order-violation penalty [38] used as a similarity, $t'$ and $v'$ denote the representations of the negative samples. Here, $[x]_+$ represents $\max(x, 0)$.

Considering we have two branches of cross-modal feature embedding, which result in two pairs of cross-modal features: the abstract features $(t_h, v_h)$ and the grounded features $(t_l, v_l)$, we modify the ranking loss as

$$\mathcal{L}_{\text{R}^+} = \sum_{t'} [\alpha - s^*(t_{h,l}, v_{h,l}) + s^*(t'_{h,l}, v_{h,l})]_+ + \sum_{v'} [\alpha - s^*(t_{h,l}, v_{h,l}) + s^*(t_{h,l}, v'_{h,l})]_+ \quad (3)$$

where $s^*(t_{h,l}, v_{h,j}) = \lambda s(t_h, v_h) + (1-\lambda) s(t_l, v_l)$ is a combined score with $\lambda$ being the tradeoff weight.

### 3.3. Image-to-text Generative Feature Learning

For the image-to-text training path (i2t, blue path in Figure 2), our goal is to encourage the *grounded* visual feature $v_l$ to be able to generate sentences that are similar to the ground-truth captions. In particular, we first encode the image with $\text{CNN}_\text{Enc}$, and then decode the grounded visual feature into a sentence with $\text{RNN}_\text{Dec}$. Like the traditional RNN-based text generation models, we first train our model on a cross-entropy (XE) loss defined as:

$$\mathcal{L}_\text{xe} = -\sum_{t=0}^{T-1} \log p_{\theta_t}(w_t|w_{0:t-1}, v_l; \theta_t) \quad (4)$$

where $w_t$ is the ground-truth word, $p_{\theta_t}(w_t|w_{0:t-1}, v_l)$ is the output probability of word $w_t$ given by the decoder with parameter $\theta_t$.

However, the XE loss is a word-level cost, and models trained on this suffer from the exposure bias problem [1] and the loss-evaluation mismatch problem [6, 31]. Thus we further employ a loss that takes the entire sequence into account. Specifically, to directly optimize the sentence-level metrics, we optimize our model by minimizing the negative expected reward given by:

$$\mathcal{L}_\text{rl} = -\mathbb{E}_{\tilde{c} \sim p_{\theta_t}}[r(\tilde{c})] \quad (5)$$

where $\tilde{c} = (\tilde{w}_0, \cdots, \tilde{w}_{T-1})$ is the word sequence sampled from the decoder, $r(\tilde{c})$ is the reward calculated by comparing the generated sentence with the corresponding reference sentences using a standard evaluation metric like BLEU [26] or CIDEr [37]. Following the reinforcement learning (RL) approach described in [33, 6], the expected gradients of Equation (5) using Monte-Carlo sample $\tilde{c}$ from $p_{\theta_t}$ can be approximated as:

$$\begin{aligned}\nabla_{\theta_t}\mathcal{L}_\text{rl} &= -\mathbb{E}_{\tilde{c} \sim p_{\theta_t}}[r(\tilde{c}) \cdot \nabla_{\theta_t} \log p_{\theta_t}(\tilde{c})] \\ &\approx -r(\tilde{c})\nabla_{\theta_t} \log p_{\theta_t}(\tilde{c}) \\ &\approx -(r(\tilde{c}) - r_b)\nabla_{\theta_t} \log p_{\theta_t}(\tilde{c})\end{aligned} \quad (6)$$

where $r_b$ is the baseline estimator used to reduce the variance without changing the expected gradient. In our model, we use the inference process reward as the baseline.

During the early stage of training, optimization of Equation (6) alone does not ensure the readability and fluency of the generated caption [27]. To deal with this, we use a mixture of XE and RL losses:

$$\mathcal{L}_\text{xe+rl} = (1-\gamma)\mathcal{L}_\text{xe} + \gamma \mathcal{L}_\text{rl} \quad (7)$$

where $\gamma$ is a tuning parameter used to balance the two losses. Equation (7) improves results on the metric used to compute the reward through the reinforcement loss but also ensures better readability and fluency due to the XE loss. For annealing and faster convergence, we start with optimizing XE loss in Equation (4), and then move to optimizing the joint loss in Equation (7).

### 3.4. Text-to-image Generative Adversarial Feature Learning

For the text-to-image training path (t2i, green path in Figure 2), our goal is to encourage the grounded text feature $t_l$ to be able to generate an image that is similar to the ground-truth one. However, unlike the image-to-text path in Section 3.3, where the model is trained to predict the word conditioned on image and history words, the reverse path suffers from the highly multi-modal distribution of images conditioned on a text representation.

The natural way to model such a conditional distribution is to use a conditional GAN [23, 32], which consists of a discriminator and a generator. The discriminator is trained to distinguish the real samples ⟨real image, true caption⟩ from the generated samples of ⟨fake image, true caption⟩ as well as samples of ⟨real image, wrong caption⟩. Specifically, the discriminator $D_i$ and the generator $G_i$ ($\text{CNN}_\text{Dec}$ in Figure 2) play the min-max game on the following value function $V(D_i, G_i)$:

$$\min_{G_i} \max_{D_i} V(D_i, G_i) = \mathcal{L}_{D_i} + \mathcal{L}_{G_i}. \quad (8)$$

The discriminator loss $\mathcal{L}_{D_i}$ and the generator loss $\mathcal{L}_{G_i}$ are defined as:

$$\begin{aligned}\mathcal{L}_{D_i} &= \mathbb{E}_{i \sim p_\text{data}}[\log D_i(i, t_l)] + \beta_f \mathbb{E}_{\hat{i} \sim p_G}[\log(1 - D_i(\hat{i}, t_l))] + \\ &\quad \beta_w \mathbb{E}_{i \sim p_\text{data}}[\log(1 - D_i(i, t'_l))]\end{aligned} \quad (9)$$

$$\mathcal{L}_{G_i} = \mathbb{E}_{\hat{i} \sim p_G}[\log(1 - D_i(\hat{i}, t_l))] \quad (10)$$

where $t_l$ and $t'_l$ denote the encoded grounded feature vectors for a matched and a mismatched captions, respectively, $i$ is the matched real image from the true data distribution $p_{\text{data}}$, $\beta_f$ and $\beta_w$ are the tuning parameters, and $\hat{i} = G_i(z, t_l)$ is the generated image by the generator $G_i$ conditioned on $t_l$ and a noise sample $z$. The variable $z$ is sampled from a fixed distribution (*e.g.*, uniform or Gaussian distribution). In implementation, we compress $t_l$ to a lower dimension and then combine it with $z$.

However, directly combining $t_l$ with $z$ cannot produce satisfactory results. This is because of the limited amount of data and the unsmoothness between $t_l$ and $z$. Thus, we introduce another variable $t_c$, which is sampled from a Gaussian distribution of $\mathcal{N}(\mu(\varphi(t_l)), \sigma(\varphi(t_l)))$ [40], where $\mu(\varphi(t_l))$ and $\sigma(\varphi(t_l))$ are the mean and the standard deviation of $t_l$, $\varphi(t_l)$ compresses $t_l$ to a lower dimension. We now generate the image conditioned on $z$ and $t_c$ with $\hat{i} = G_i(z, t_c)$. The discriminator loss $\mathcal{L}_{D_i}$ and the generator loss $\mathcal{L}_{G_i}$ are then modified to:

$$\mathcal{L}_{D_i} = \mathbb{E}_{i \sim p_{\text{data}}}[\log D_i(i, t_l)] + \beta_f \mathbb{E}_{\hat{i} \sim p_G}[\log(1 - D_i(\hat{i}, t_l))] + \beta_w \mathbb{E}_{i \sim p_{\text{data}}}[\log(1 - D_i(i, t'_l))] \quad (11)$$

$$\mathcal{L}_{G_i} = \mathbb{E}_{\hat{i} \sim p_G}[\log(1 - D_i(\hat{i}, t_l))] + \beta_s \mathcal{D}_{\text{KL}}(\mathcal{N}(\mu(\varphi(t_l)), \sigma(\varphi(t_l))) \parallel \mathcal{N}(0, 1)) \quad (12)$$

where $\beta_f$, $\beta_w$ and $\beta_s$ are the tuning parameters, and the KL-divergence term is to enforce the smoothness of the latent data manifold.

Alg. 1 summarizes the entire training procedure.

## 4. Experiments

### 4.1. Dataset and Implementation Details

We evaluate our approach on the MSCOCO dataset [19]. For cross-modal retrieval, we use the setting of [12], which contains 113,287 training images with five captions each, 5,000 images for validation and 5,000 images for testing. We experiment with two image encoders: VGG19 [35] and ResNet152 [9]. For VGG19, we extract the features from the penultimate fully connected layer. For ResNet152, we obtain the global image feature by taking a mean-pooling over the last spatial image features. The dimensions of the image feature vectors is 4096 for VGG19 and 2048 for ResNet152. As for text preprocessing, we convert all sentences to lower case, resulting in a vocabulary of 27,012 words.

We set the word embedding size to 300 and the dimensionality of the joint embedding space to 1024. For the sentence encoders, we use a bi-directional GRU-based encoder to get the *abstract* feature representation $t_h$ and one GRU-based encoder to get the *grounded* feature representation $t_l$. The number of hidden units of both GRUs is set to 1024. For the sentence decoder, we adopt a one-layer GRU-based decoder which has the same hidden dimensions as the

---

**Algorithm 1** GXN training procedure.

**Input:** Positive image $i$, negative image $i'$, positive text $c$, negative text $c'$, number of training batch steps $S$
1: **for** $n = 1 : S$ **do**
2:     /\*Look\*/
3:     Draw image-caption pairs: $(i, c)$, $i'$ and $c'$.
4:     $v_h, v_l, v'_h, v'_l \leftarrow i, i'$ {Image encoding}
5:     $t_h, t_l, t'_h, t'_l \leftarrow c, c'$ {Text encoding}
6:     Update parameters with Gen$_{\text{i2t}}$-GXN
7:     Update parameters with Gen$_{\text{t2i}}$-GXN
8: **end for**

**Function:** Gen$_{\text{i2t}}$-GXN
1: /\*Imagine\*/
2: $\hat{c} = \text{RNN}_{\text{Dec}}(v_l, c)$ {Scheduled sampling}
3: Compute XE loss $\mathcal{L}_{\text{xe}}$ using (4).
4: $\tilde{c} \leftarrow \text{RNN}_{\text{Dec}}(v_l)$ {Sampling}
5: $\bar{c} \leftarrow \text{RNN}_{\text{Dec}}(v_l)$ {Greedy decoding}
6: Compute RL loss $\mathcal{L}_{\text{rl}}$ using (5).
7: Update model parameters by descending stochastic gradient of (7) with $r_b = r(\bar{c})$ (see (6)).
8: /\*Match\*/
9: Update model parameters using (3).

**Function:** Gen$_{\text{t2i}}$-GXN
1: /\*Imagine\*/
2: $t_c \sim \mathcal{N}(\mu(\varphi(t_l)), \sigma(\varphi(t_l)))$
3: $\hat{i} = G_i(z, t_c)$
4: Update image discriminator $D_i$ using (11).
5: Update image generator $G_i$ using (12).
6: /\*Match\*/
7: Update model parameters using (3).

---

two GRU-based encoders. During the RL training, we use CIDEr score as the sentence-level reward. We set $\beta_f = 0.5$, $\beta_w = 0.5$ and $\beta_s = 2.0$ in Eq. (11) and (12), margin $\alpha$ and $\lambda$ in Eq. (3) to be 0.05 and 0.5 respectively, and $\gamma$ in Eq. (7) is increased gradually based on the epoch from 0.05 to 0.95. The output size of the image decoder CNN$_{\text{Dec}}$ is $64 \times 64 \times 3$, and the real image is resized before inputting to the discriminator. All the modules are randomly initialized before training except for the CNN encoder and decoder. Dropout and batch normalization are used in all our experiments. We use Adam [14] for optimization with a mini-batch size of 128 in all our experiments. The initial learning rate is 0.0002, and the momentum is 0.9.

For evaluation, we use the same measures as those in [38], *i.e.*, R@K, defined as the percentage of queries in which the ground-truth matchings are contained in the first K retrieved results. The higher value of R@K means better performance. Another metric we use is Med $r$, which is the median rank of the first retrieved ground-truth sentence or image. The lower its value, the better. We also compute another score, denoted as '**Sum**', to evaluate the overall per-

formance for cross-modal retrieval, which is the summation of all R@1 and R@10 scores defined as follows:

$$\text{Sum} = \underbrace{R@1 + R@10}_{\text{Image-to-Text}} + \underbrace{R@1 + R@10}_{\text{Text-to-Image}} \quad (13)$$

In addition, we evaluate the quality of the generated captions with the standard evaluation metrics: CIDEr and BLEU-$n$. BLEU-$n$ rates the quality of the retrieved captions by comparing $n$-grams of the candidate with the $n$-grams of the five gold captions and count the number of matches. CIDEr is a consensus-based metric which is more correlated with human assessment of caption quality.

### 4.2. Baseline Approaches for Comparisons

**GRU (VGG19) and GRU$_{\text{Bi}}$ (VGG19):** These two baselines use the pre-trained VGG19 as the image encoder. GRU (VGG19) adopts a one layer GRU as the sentence encoder, while GRU$_{\text{Bi}}$ (VGG19) adopts a bi-directional GRU as the sentence encoder. These two models are trained using Eq. (2).

**GXN (ResNet152) and GXN (fine-tune):** These two baselines use the same two GRU sentence encoders as our proposed GXN framework, but without the generation components. In other words, they only contain the cross-modal feature embedding training path using Eq. (3). Here, the pre-trained ResNet152 is adopted as the image encoder. GXN (ResNet152) and GXN (fine-tune) refer to the models without or with fine-tuning ResNet152, respectively. The fine-tuned ResNet152 model is used as the image encoder for all other GXN models.

**GXN (i2t, xe) and GXN (i2t, mix):** These two GXN baseline models contain not only the cross-modal feature embedding training path but also the image-to-text generative training path. GXN (i2t, xe) and GXN (i2t, mix) are the two models optimized with Eq. (4) and (7), respectively.

**GXN (t2i):** This baseline model contains both the cross-modal feature embedding training path and the text-to-image generative training path, and is trained with Gen$_{\text{t2i}}$-GXN in Algorithm 1.

**GXN (i2t+t2i):** This is our proposed full GXN model containing all the three training paths. It is initialized with the trained parameters from GXN (i2t, mix) and GXN (t2i) and fine-tuned with Algorithm 1.

### 4.3. Quantitative Results

In this section, we present our quantitative results and analysis. To verify the effectiveness of our approach and to analyze the contribution of each component, we compare different baselines in Table 1 and 2. The comparison of our approach with the state-of-the-art methods is shown in Table 3.

**Effect of a Better Text Encoder.** The first two rows in Table 1 compare the effectiveness of the two sentence encoders. Compared with GRU (VGG19), GRU$_{\text{Bi}}$ (VGG19) can make full use of the context information from both directions and achieve better performance, *i.e.*, GRU$_{\text{Bi}}$ (VGG19) increases the caption retrieval R@1 from 51.4 to 53.6 and image retrieval R@1 from 39.1 to 40.0.

Table 1: Cross-modal retrieval results on MSCOCO 1K-image test set (bold numbers are the best results).

| Model | Image-to-Text | | | Text-to-Image | | |
|---|---|---|---|---|---|---|
| | R@1 | R@10 | Med | R@1 | R@10 | Med |
| GRU(VGG19) | 51.4 | 91.4 | **1.0** | 39.1 | 86.7 | 2.0 |
| GRU$_{\text{Bi}}$(VGG19) | 53.6 | 90.2 | **1.0** | 40.0 | 87.8 | 2.0 |
| GXN(ResNet152) | 59.4 | 94.7 | **1.0** | 47.0 | 92.6 | 2.0 |
| GXN(fine-tune) | 64.0 | 97.1 | **1.0** | 53.6 | 94.4 | **1.0** |
| GXN(i2t,xe) | 68.2 | 98.0 | **1.0** | 54.5 | **94.8** | **1.0** |
| GXN(i2t,mix) | 68.4 | 98.1 | **1.0** | 55.6 | 94.6 | **1.0** |
| GXN(t2i) | 67.1 | **98.3** | **1.0** | 56.5 | **94.8** | **1.0** |
| GXN (i2t+t2i) | **68.5** | 97.9 | **1.0** | **56.6** | 94.5 | **1.0** |

**Effect of a Better Image Encoder.** We further investigate the effect of image encoding model on the cross-modal feature embedding. By replacing the VGG19 model in GRU$_{\text{Bi}}$ (VGG19) with ResNet152, we achieve huge performance gains. The caption retrieval R@1 increases from 53.6 to 64.0, and the image retrieval R@1 increases from 40.0 to 53.6.

**Effect of the Generative Models.** We first consider the incorporation of the image-to-caption generation process into our GXN model. From Table 1, it can be seen that, compared with GXN (fine-tune), GXN (i2t, xe) achieves significantly better performance on the image-to-text retrieval. This validates our assumption that by combining the *abstract* representation with the *grounded* representation learned by caption generation (imagining), we can retrieve more relevant captions. Then, as we further enrich the model with the mixed RL+XE loss of Eq. (7), we observe further improvements (see GXN (i2t, mix)).

We also evaluate the effect of incorporating the text-to-image generation process into our GXN model. It can be seen from Table 1 that, compared with GXN (fine-tune), GXN (t2i) significantly improves the text-to-image retrieval performance. This is because the *grounded* text feature $t_l$ is well learned via the text-to-image generation process (imagining). Although the image-to-text retrieval performance of GXN (t2i) is not as good as GXN (i2t, mix), it is still much better than GXN (fine-tune), which does not incorporate any generative process.

The final row in Table 1 shows the performance of our complete model, *i.e.*, GXN (i2t+t2i), which incorporates both image and text generations. We can see that GXN (i2t+t2i) achieves the best performances in general, having the advantages of both GXN (i2t, mix) and GXN (t2i).

**Quality of the retrieved captions.** For the image-to-text retrieval task in Table 1, Table 2 reports the quality of the retrieved captions using the sentence-level metrics, BLEU and CIDEr. Both BLEU and CIDEr have been shown to correlate well with human judgments [37]. As shown in Table 2, incorporating the generative models into GXN yields better results than GXN (fine-tune) that does not incorporate any generation process. Note that those scores are calculated over five reference sentences. This demonstrates that our proposed GXN model can retrieve captions that are closer to the ground-truth ones.

Table 2: Evaluating the quality of the retrieved captions on MSCOCO 1K test set using the sentence-level metrics, where B@n is a short form for BLEU-$n$, and C is a short form for CIDEr. All values are reported in percentage. The 2nd column is the rank order of the retrieved caption.

| Model | No. | B@1 | B@2 | B@3 | B@4 | C |
|---|---|---|---|---|---|---|
| GXN(fine-tune) | 1 | 54.6 | 34.5 | 21.0 | 12.9 | 56.3 |
| GXN(i2t,xe) | 1 | 56.5 | 36.2 | 22.6 | 14.1 | 59.2 |
| GXN(i2t,mix) | 1 | 57.0 | 36.7 | 23.0 | 14.4 | 60.0 |
| GXN(t2i) | 1 | 56.0 | 36.0 | 22.4 | 14.3 | 58.8 |
| GXN(t2i+t2i) | 1 | **57.1** | **36.9** | **23.3** | **14.9** | **61.1** |
|  | 2 | 55.8 | 35.8 | 22.4 | 13.7 | 58.3 |
|  | 3 | 54.2 | 33.6 | 20.5 | 12.7 | 54.0 |
|  | 4 | 53.1 | 32.9 | 19.9 | 11.9 | 51.2 |
|  | 5 | 53.2 | 32.8 | 19.6 | 11.3 | 51.1 |

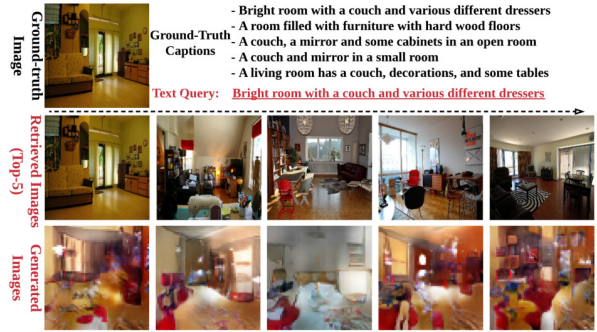

Figure 4: Visual results of text-to-image retrieval. 2nd row: retrieved images. 3rd row: image samples generated by our conditional GAN.

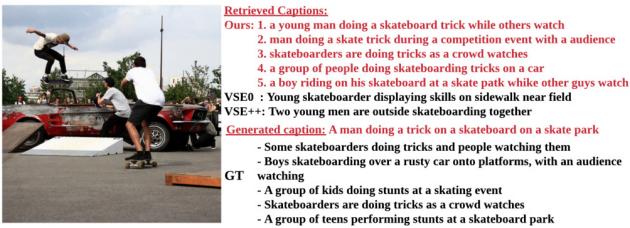

Figure 3: Visual results of image-to-text retrieval, where the top-5 retrieved captions and the generated caption are shown in red color.

### 4.3.1 Comparisons with the State-of-the-art

Table 3 shows the comparisons of our cross-modal retrieval results on MSCOCO dataset with state-of-the-art methods. We can see that our framework achieves the best performance in all metrics, which clearly demonstrates the advantages of our model. To make our approach more convincing and generic, we also conduct experiments on Flickr30K dataset with results shown in Table 4.

### 4.4. Qualitative Results

In this section, we present a qualitative analysis of our GXN (i2t+t2i) framework on cross-modal retrieval.
**Results of image-to-text retrieval.** Figure 3 depicts some examples for image-to-text retrieval, where the results of VSE0 and VSE++ are adopted from [4]. We show the top-5 retrieved captions as well as the ground-truth captions. We can see that the retrieved captions of our model can better describe the query images.

**Results of text-to-image retrieval.** Figure 4 depicts some examples for text-to-image retrieval, where we show the top-5 retrieved images as well as the generated images. Compared to the ground-truth image and the retrieved images, although the generated images are of limited quality for complex multi-object scenes, they still contain certain plausible shapes, colors, and backgrounds. This suggests that our model can capture the complex underlying language-image relations.

Some more samples are shown in Figure 5. We show the retrieved and generated results for both image-to-text and text-to-image on the same image-caption pairs.
**Results of word embedding.** As a byproduct, a word embedding matrix $\mathbf{W}_e$ (mentioned at the beginning of Section 3.2) is also learned in our GXN models. We visualize the learned word embedding by projecting some selected word vectors into a 2-D space in Figure 6. We can see that compared with the embeddings learned from GXN (fine-tune), our GXN (i2t+t2i) can learn word embedding with more related visual meaning. For example, we find that words like '*eats*' and '*stares*' of GXN (i2t+t2i) are closer to each other compared to those of GXN (fine-tune). This is also consistent with the fact that when we '*eat*' some food; we also tend to '*stare*' at it.

## 5. Conclusion

In this paper, we have proposed a novel cross-modal feature embedding framework for cross image-text retrieval. The uniqueness of our framework is that we incorporate the image-to-text and the text-to-image generative models into the conventional cross-modal feature embedding. We learn both the high-level abstract representation and the lo-

Table 3: Comparisons of the cross-modal retrieval results on MSCOCO dataset with the state-of-the-art methods. We mark the unpublished work with ∗ symbol. Note that 'Sum' is the summation of the two R@1 scores and the two R@10 scores.

| Model | Image-to-Text Retrieval | | | Text-to-Image Retrieval | | | Sum |
|---|---|---|---|---|---|---|---|
| | R@1 | R@10 | Med $r$ | R@1 | R@10 | Med $r$ | |
| | *1K Test Images* | | | | | | |
| m-CNN [20] | 42.8 | 84.1 | 2.0 | 32.6 | 82.8 | 3.0 | 242.3 |
| HM-LSTM [25] | 43.9 | 87.8 | 2.0 | 36.1 | 86.7 | 3.0 | 254.5 |
| Order-embeddings [38] | 46.7 | 88.9 | 2.0 | 38.9 | 85.9 | 2.0 | 260.4 |
| DSPE+Fisher Vector [39] | 50.1 | 89.2 | - | 39.6 | 86.9 | - | 265.8 |
| sm-LSTM [10] | 53.2 | 91.5 | **1.0** | 40.7 | 87.4 | 2.0 | 272.8 |
| ∗VSE++ (ResNet152, fine-tune) [4] | 64.7 | 95.9 | **1.0** | 52.0 | 92.0 | **1.0** | 304.6 |
| GXN (i2t+t2i) | **68.5** | **97.9** | **1.0** | **56.6** | **94.5** | **1.0** | **317.5** |
| | *5K Test Images* | | | | | | |
| Order-embeddings [38] | 23.3 | 65.0 | 5.0 | 18.0 | 57.6 | 7.0 | 163.9 |
| ∗VSE++ (ResNet152, fine-tune) [4] | 41.3 | 81.2 | **2.0** | 30.3 | 72.4 | 4.0 | 225.2 |
| GXN(t2i+t2i) | **42.0** | **84.7** | **2.0** | **31.7** | **74.6** | 3.0 | **233.0** |

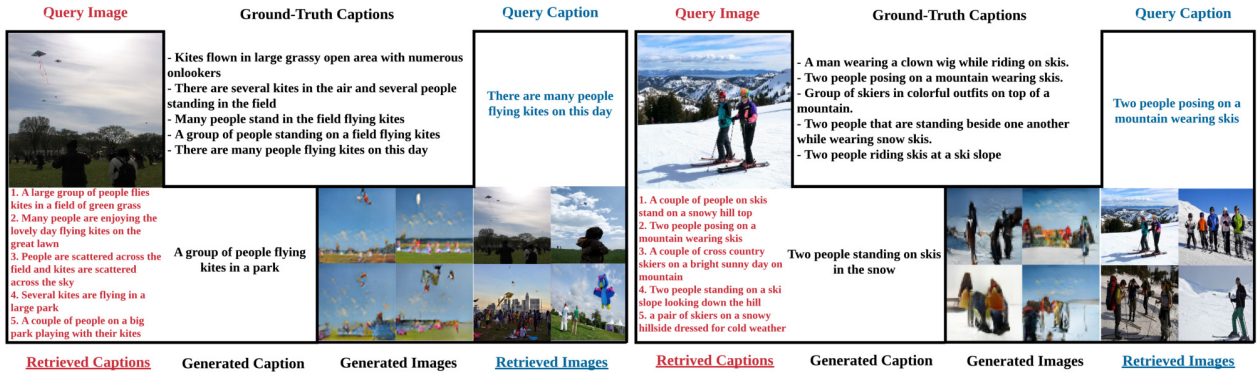

Figure 5: More visual results of cross-modal retrieval.

Table 4: Experimental results on Flickr30K 1k image test set.

| Model | Image-to-Text | | | Text-to-Image | | |
|---|---|---|---|---|---|---|
| | R@1 | R@10 | Med $r$ | R@1 | R@10 | Med $r$ |
| ∗VSE++ [4] | 52.9 | 87.2 | **1.0** | 39.6 | 79.5 | **2.0** |
| GXN (i2t+t2i) | **56.8** | **89.6** | **1.0** | **41.5** | **80.1** | **2.0** |

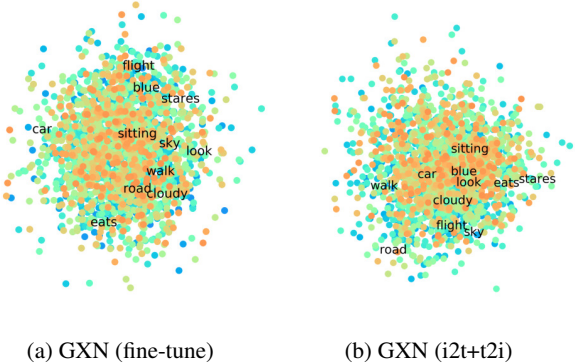

(a) GXN (fine-tune)  (b) GXN (i2t+t2i)

Figure 6: Visualization of word embedding.

cal grounded representation of multi-modal data in a max-margin learning-to-rank framework. Our framework significantly outperforms state-of-the-art methods for textural-visual cross-modal retrieval on MSCOCO dataset. Future research directions include considering pixel-level visual quality assessment and other strong discriminators to improve the quality of the generated images.

## Acknowledgments

This research was supported by the National Research Foundation, Prime Minister's Office, Singapore, under its IDM Futures Funding Initiative, and NTU CoE Grant. This research was carried out at the Rapid-Rich Object Search (ROSE) Lab at the Nanyang Technological University, Singapore. The ROSE Lab is supported by the National Research Foundation, Singapore, and the Infocomm Media Development Authority, Singapore. We gratefully acknowledge the support of NVAITC (NVIDIA AI Tech Center) for our research at NTU ROSE Lab, Singapore.